\documentclass[runningheads]{llncs}
\usepackage{graphicx}

\usepackage{epsfig}
\usepackage{amsmath}
\usepackage{amssymb}

\usepackage{booktabs}
\usepackage{diagbox}
\usepackage{stackengine}

\usepackage{color}
\usepackage{xcolor}
\usepackage{colortbl}
\definecolor{maroon}{cmyk}{0,0.87,0.68,0.32}

\usepackage{threeparttable}
\usepackage{multirow}
\begin{document}

\title{MMSummary: Multimodal Summary Generation for Fetal Ultrasound Video}

\author{Xiaoqing Guo\inst{1}\orcidID{0000-0002-9476-521X} \and
Qianhui Men\inst{1}\orcidID{0000-0002-0059-5484} \and ~~
J. Alison Noble\inst{1}\orcidID{0000-0002-3060-3772}}
% index{Men, Qianhui}
% index{Guo, Xiaoqing}
% index{Noble, Alison}

\authorrunning{Xiaoqing Guo, Qianhui Men, and J. Alison Noble}
 
\institute{Department of Engineering Science, University of Oxford, Oxford, UK
\email{xiaoqing.guo@eng.ox.ac.uk}}

\maketitle 

\begin{abstract}

We present the first automated multimodal summary generation system, MMSummary, for medical imaging video, particularly with a focus on fetal ultrasound analysis. Imitating the examination process performed by a human sonographer, MMSummary is designed as a three-stage pipeline, progressing from \textit{keyframe detection} to \textit{keyframe captioning} and finally anatomy \textit{segmentation and measurement}. In the \textit{keyframe detection} stage, an innovative automated workflow is proposed to progressively select a concise set of keyframes, preserving sufficient video information without redundancy. Subsequently, we adapt a large language model to generate meaningful captions for fetal ultrasound keyframes in the \textit{keyframe captioning} stage. If a keyframe is captioned as fetal biometry, the \textit{segmentation and measurement} stage estimates biometric parameters by segmenting the region of interest according to the textual prior. The MMSummary system provides comprehensive summaries for fetal ultrasound examinations and based on reported experiments is estimated to reduce scanning time by approximately 31.5\%, thereby suggesting the potential to enhance clinical workflow efficiency.


\end{abstract}

\section{Introduction}

\begin{figure}[t]
\centering
\includegraphics[width=1.\textwidth]{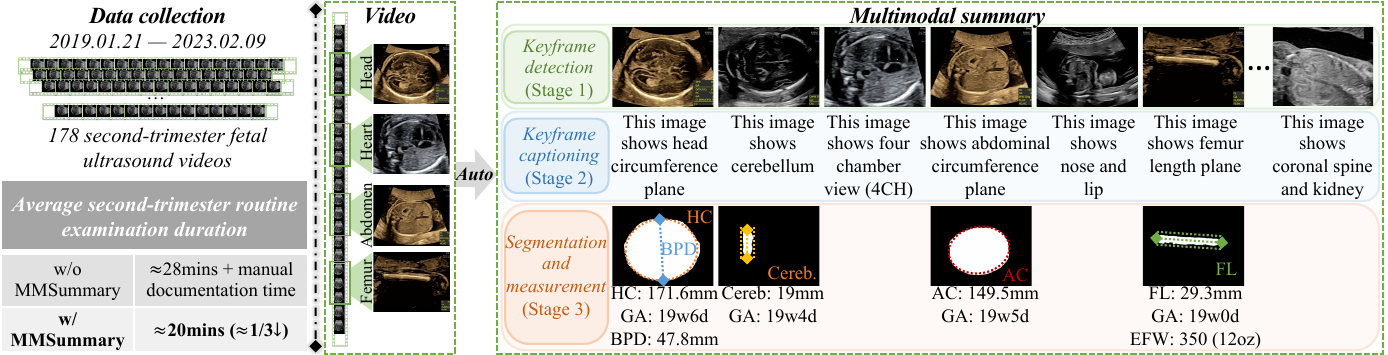}
\caption{Multimodal summary for fetal US video to support sonography workflow.} \label{fig:introduction}
\end{figure}

Ultrasound (US) examinations are essential for monitoring fetal development and maternal health throughout pregnancy. Conducting these examinations requires skill to carefully \textit{manipulate the US probe to locate anatomies, interpret images, and perform biometry} \cite{selvathi2022fetal}. However, learning to scan well is hard and can take years of training \cite{le2021impact}. As a result, there is a worldwide shortage of highly-skilled qualified sonographers, including in low-and-middle-income countries, point-of-care environments, and emergency medical situations \cite{self2022developing}. 
Fetal screening is also time-consuming \cite{sharma2021knowledge}, often taking around 28 minutes per second-trimester examination, as observed in our dataset (left of Fig. \ref{fig:introduction}). Moreover, the redundancy in US videos complicates retrospective analysis and documentation. These challenges necessitate an automated summarizing system capable of efficiently highlighting useful content in examinations and ensuring accurate assessments regardless of expertise, even when examinations need to be conducted quickly.

Such an automated summarizing system differs from video summarization in computer vision \cite{apostolidis2021video}, posing its unique challenges. First, US videos exhibit frame redundancy in appearance, and frames of the same anatomy may have different appearances due to the probe positioning. Hence, it is crucial to detect a subset of representative keyframes rather than potentially redundant video clips as in traditional video summarization. Second, this system should not only identify useful content from the video, but also interpret anatomies and findings within the identified content, and even measure important parameters (right of Fig. \ref{fig:introduction}).

In this paper, we propose {MMSummary}, a multimodal summary generation system for medical imaging video, with a particular focus on fetal US. Mimicking the examination process, our method comprises a three-stage pipeline, consisting of {\textit{keyframe detection}}, {\textit{keyframe captioning}}, and {\textit{segmentation and measurement}}. In the \textbf{\textit{keyframe detection}} stage, a transformer-based network mines temporal information within the video to detect representative frames with anatomical structures of interest. Then a diverse keyframe detection algorithm is designed to eliminate redundant keyframes, forming a concise subset of keyframes. 
In the \textbf{\textit{keyframe captioning}} stage, following ClipCap \cite{mokady2021clipcap}, we leverage pre-trained foundation models, namely BiomedCLIP~\cite{zhang2023large} and GPT2~\cite{radford2019language}, which possess a broad understanding of biomedical visual and textual data. By only fine-tuning a lightweight mapping network to bridge the gap between visual and textual features, the large language model GPT2 is adapted to generate meaningful and user-friendly captions for detected keyframes. 
If a keyframe is captioned as fetal biometry, the \textbf{\textit{segmentation and measurement}} stage segments a region of interest for automated biometry. Unlike existing fetal anatomy segmentation methods \cite{bano2021autofb,zhao2023transfsm,nagabotu2023precise}, we incorporate prior textual information as prompts to guide anatomy segmentation. 
By automating the process of US video summarization, we show in our experiments that MMSummary saves around 31.5\% scanning time, expedites and simplifies the scan documentation process, and can achieve consistent and accurate assessments independent of operator expertise.

\textbf{Related Work:} (1) Automated  \textbf{\textit{medical report/summary generation}} has been explored in several medical image analysis applications including for chest X-ray images \cite{wang2023chatcad,yang2023radiology,zhou2021visual}, 
breast US images \cite{ge2023ai,huh2023breast,yang2021automatic}, and fetal US images \cite{alsharid2022weakly}. 
Typically, it is framed as a visual captioning task, where medical image features extracted from convolutional neural networks are input into recurrent neural networks to produce word sequences, forming medical reports, and the recurrent neural network is usually trained from scratch \cite{alsharid2022weakly,yang2021automatic,zhou2021visual}. The recent emergence of large language models has led to significant advancement in report generation \cite{huh2023breast,wang2023chatcad}. 
These methods are primarily tailored to 2D medical images and text-based summarization, except \cite{tiwari2023experience} which provides multimodal summarization of clinical conversation. To the best of our knowledge, we are the first to generate multimodal summaries for medical imaging video. 
(2) \textbf{\textit{Video summarization}} is extensively studied in computer vision, aiming to generate a concise representation of video content defined as a set of keyframes or clips \cite{apostolidis2021video}. Early non-deep learning-based methods focused on selecting keyframes \cite{de2011vsumm,furini2010stimo}. Recent methods prioritize video clips \cite{dvornik2023stepformer,li2023progressive,lin2023univtg,wu2022intentvizor,zala2023hierarchical} since the clips incorporate audio and motion for a more natural narrative and enriched summarization. In fetal US video, one prior work employed reinforcement learning to extract video clips \cite{liu2020ultrasound}. However, frames within clips may be too numerous to include in a multimodal summary. Moreover, and importantly, in fetal US video, keyframes often exhibit a similar anatomical appearance to neighboring frames, leading to frame redundancy. Thus, our approach focuses on extracting keyframes sufficient to summarize the whole examination with minimal frame redundancy. Additionally, unlike prior video summarization methods, our MMSummary generates text descriptions for the extracted keyframes.

\section{Method}

\begin{figure}[t]
\centering
\includegraphics[width=1.\textwidth]{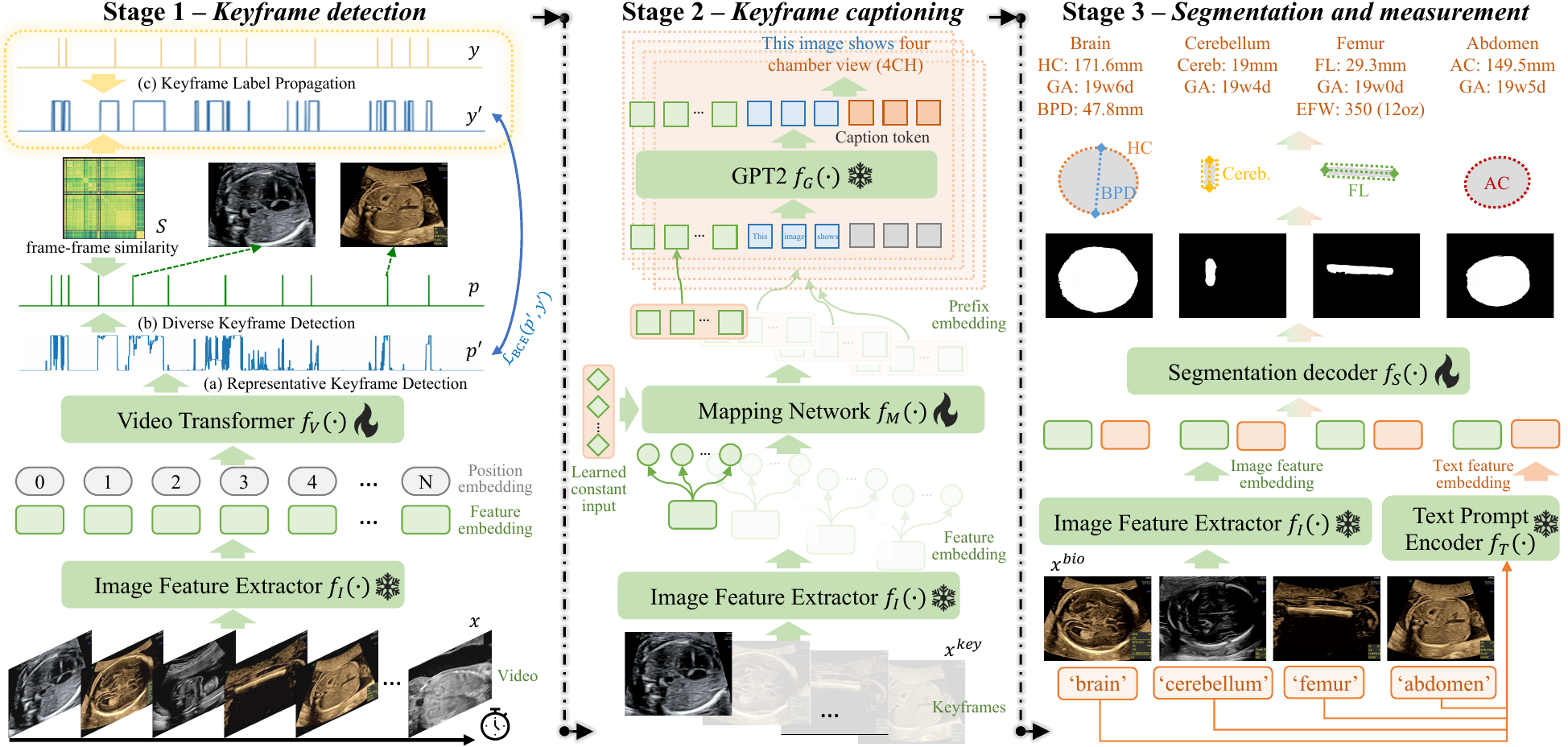}
\caption{Illustration of the three-stage MMSummary pipeline.} \label{fig:framework}
\end{figure}

The proposed three-stage MMSummary pipeline is illustrated in Fig. \ref{fig:framework}. Given an input video, MMSummary generates a multimodal summarization with keyframes, text descriptions, and biometry parameters. In \textbf{\textit{keyframe detection}} (stage 1), the method takes the video as input $\mathcal{V} = \{ x_{1}, ..., x_{T} \}$, which spans over $T$ time steps. The keyframe label for the input video is denoted as $y=\{ y_{1}, ..., y_{T}\}$, where $y_{t}=1$ indicates a keyframe, and otherwise, it is a non-keyframe. Keyframes extracted in stage 1 are separately fed into stage 2 for \textbf{\textit{keyframe captioning}}. Among these keyframes, $\{ x^{key}_{1}, ..., x^{key}_{k} \}$ are annotated with caption labels $\{ y^{key}_{1}, ..., y^{key}_{k} \}$. Additionally, some keyframes $\{ x^{bio}_{1}, ..., x^{bio}_{b} \}$ can be used for biometry estimation and are equipped with region-of-interest mask labels $\{ y^{bio}_{1}, ..., y^{bio}_{b} \}$. In \textbf{\textit{segmentation and measurement}} (stage 3), a unified model is learned that automatically segments regions of interest, and estimates biometric parameters, gestational age, and fetal weight. 

\subsubsection{\textit{Keyframe detection.}} This stage detects a concise subset of representative and diverse keyframes that are distinct from each other, summarizing the useful anatomical content in the video. As in Figure \ref{fig:framework} (left side), frames of the video are passed through a pre-trained image feature extractor ($f_{I}(\cdot)$, BiomedCLIP ViT-B \cite{zhang2023large}) followed by a learnable linear layer, yielding a sequence of feature embeddings. These embeddings are added to vanilla sine-cosine temporal position embeddings, and fed into a 4-layer video transformer $f_{V}(\cdot)$ (each containing a multi-head self-attention and MLP), outputting the probability of each frame being a keyframe $p'\in [0,1]^{T}$. To optimize the learnable linear layer and video transformer, binary cross-entropy loss $\mathcal{L}_{BCE}(p', y)=-y\cdot log (p') - (1-y)\cdot log(1-p')$ is computed w.r.t. the keyframe label $y$. 

When minimizing $\mathcal{L}_{BCE}(p', y)$, we encounter extreme class imbalance in our application with the keyframe label $y$ exhibiting a class ratio over 1600:1. Indeed, due to fine-grained probe adjustments, image annotation, and biometry measurements performed by sonographers, keyframe neighbors often share a similar feature representation. Based on this observation, we propose a \textit{keyframe label propagation} algorithm to balance the non-keyframe and keyframe classes in the label distribution. Specifically, we introduce a frame-frame similarity matrix computed via the cosine similarity $S\in [0,1]^{T\times T}$ between frame features derived from \cite{baumgartner2017sononet} within the video to propagate keyframe labels. If a frame exhibits a high similarity score (above 0.99) compared with any keyframes within the video, it is relabeled as a keyframe, resulting in an updated keyframe label $y'$. Thus, by minimizing $\mathcal{L}_{BCE}(p', y')$, the model is optimized to \textit{detect representative frames} that preserve as much essential information across the entire video as possible. 

To mitigate redundancy in detected frames, we further propose an \textit{diverse keyframe detection} algorithm (\textit{supp.} Fig. \ref{fig:sup}), which iteratively selects keyframes to ensure exclusiveness in the resulting subset. In each iteration, the frame with the highest probability in the video is selected as the keyframe, and then we mask out other frames that share similar features with the selected keyframe (i.e., have a similarity above $\tau$). This process continues iteratively until the highest probability falls below $\tau'$. Finally, we obtain a concise set of keyframes.

\begin{figure}[ht!]
\centering
\includegraphics[width=0.99\textwidth]{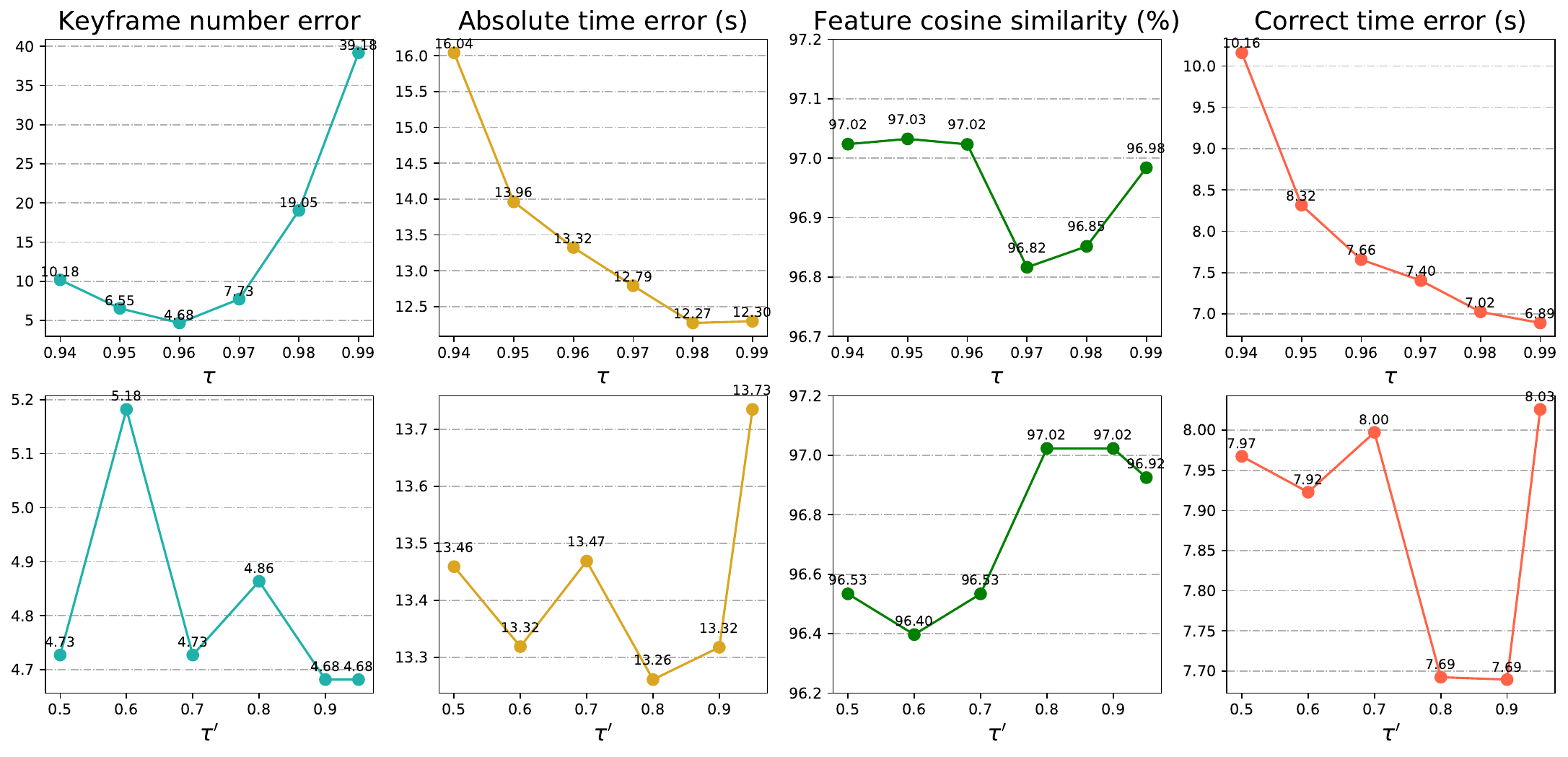}
\caption{Hyper-parameters $\tau$ and $\tau'$.} 
\label{fig:alpha}
\end{figure}

It is worth noting that choosing good values for $\tau$ and $\tau'$ is crucial  (\textit{supp.} Fig. \ref{fig:alpha}). A too large $\tau$ cannot thoroughly filter out similar frames, leading to frame redundancy. Conversely, a too small $\tau$ may result in a risk of information loss as some representative frames are discarded. 
As for $\tau'$, a too small value may lead to the inclusion of non-informative frames, while choosing a too large value may result in information loss due to discarding potentially useful frames. Hence, well chosen $\tau$ and $\tau'$ values can ensure the selected keyframes effectively capture the essential information while minimizing redundancy and information loss.

\subsubsection{\textit{Keyframe captioning.}} To automatically generate text descriptions for detected keyframes, we use BiomedCLIP (i.e., image feature extractor $f_{I}(\cdot)$) \cite{zhang2023large} and GPT2 \cite{radford2019language} ($f_{G}(\cdot)$), which possess a broad understanding of biomedical visual and textual data, respectively. Then we introduce a lightweight mapping network $f_{M}(\cdot)$ to bridge the gap between visual and textual features. It takes two inputs, visual feature embeddings and a learned constant input. The constant retrieves meaningful information from visual feature embeddings through multi-head attention, resulting in $m=50$ prefix embeddings, i.e., $p^{emb}_{1}, ..., p^{emb}_{m}=f_{M}(f_{I}(x^{key}))$. Additionally, we use the tokenizer of GPT2 to generate a sequence of embeddings for text prompt `\textit{this image shows}', i.e., $t^{emb}_{1}, t^{emb}_{2}, t^{emb}_{3}$, which, together with previous prefix embeddings, serve as input prompts for GPT2. 
 
During the training phase, GPT2 model outputs are constrained with the caption label $y^{key}$, i.e., a sequence of tokens $y^{key,1}, ..., y^{key,l}$. Our training objective is to predict caption tokens conditioned on both visual and textual prompts in an autoregressive manner. Specifically, we optimize the mapping network using the cross-entropy loss $\mathcal{L}_{CE}^{caption}=-\sum_{i=1}^{l}log p(y^{key,i} \mid p^{emb}_{1}, ...,p^{emb}_{m}, t^{emb}_{1}, ..., t^{emb}_{3},$ 
$y^{key,1}, ..., y^{key,i-1})$. By optimizing the mapping network to refine prefix embeddings, GPT2 is effectively adapted to generate meaningful and user-friendly captions for detected keyframes. 
In the inference phase, we extract the visual prefix embeddings for detected keyframes from the previous stage and obtain text prompt embeddings as well. Then we start generating a caption conditioned on these prompts by, iteratively, predicting the next token. At each step, the language model outputs probabilities for all vocabulary tokens, and we select the token with the highest probability as the next token in the caption sequence.

\subsubsection{\textit{Segmentation and measurement.}} If a keyframe $x^{key}$ is recognized as a biometry image $x^{bio}$, it is fed into the segmentation model to segment the region of interest for biometric parameter measurement. 
Given that semantic information embedded in text can enhance medical image segmentation \cite{du2023segvol}, we incorporate prior textual information generated from the \textit{keyframe captioning} stage as prompts to guide segmentation of the corresponding anatomy.

Specifically, input image $x^{bio}$ is fed into the image feature extractor (ViT-B) pre-trained from \cite{ma2024segment}, yielding a feature size of 64$\times$64 with a 256-dimensional embedding vector $f_{I}(x^{bio})$. Simultaneously, the corresponding caption is input to the text prompt encoder $f_{T}(\cdot)$ (pre-trained from CLIP \cite{radford2021learning}) followed by a trainable projection head to generate a 256-dimensional text feature embedding. Then a lightweight mask decoder, comprising two transformer layers and two convolutional layers, enhances the image embedding with text prompt features. The resulting embedding is upsampled and processed by a dynamic mask prediction head to derive the final mask foreground probability $p^{bio}$ at each image location. The projection heads and mask decoder are optimized by the combination of binary cross-entropy loss and dice loss w.r.t. the mask label $y^{bio}$.

Specific to our video analysis application, there are four biometry planes: head circumference (HC), abdominal circumference (AC), femur length (FL), and head cerebellum (Cereb). Empirically the head and abdomen exhibit elliptical shapes, while the cerebellum and femur are measured by lines. Therefore, we employ ellipse fitting on the segmented head and abdomen masks using a least-squares-fit algorithm \cite{fitzgibbon1999direct}. The perimeter of the fitted ellipse represents HC and AC respectively. For the head, the minor axis of the fitted ellipse is the biparietal diameter (BPD) estimate. For the cerebellum and femur, we fit a minimum external rectangle onto the predicted masks. The long side of the rectangle is used as the cerebellum and femur length estimate, respectively. Lengths measured in pixels are scaled to millimeters using a caliper visible on the left-hand side of the US image \cite{bano2021autofb}. This allows for direct comparison with clinically obtained measurements. Finally, the estimated biometry is converted to gestational age and estimated fetal weight using clinical equations~\cite{kiserud2017world,maraci2020toward}.

\section{Experiments}

\textbf{Dataset.} The dataset utilized in this study is from PULSE (Perception Ultrasound by Learning Sonographic Experience) \cite{drukker2021transforming}, with approval fromUK Research Ethics Committee. Clinical fetal US scans were conducted using a GE Voluson E8 scanner. The videos were captured at 30 Hz and downsampled to 5 Hz. Our study included 178 second-trimester routine examination videos conducted by 3 newly qualified (less than two years of experience) and 1 expert (14 years of experience) sonographers. These videos were randomly split into training (145), validation (11), and test (22) sets. The ground-truth keyframes containing meaningful anatomical structures were obtained from screenshots captured by sonographers during clinical scanning. This resulted in 4596, 329, 679 keyframes in the three sets, of which 3481, 254, 531 frames had caption labels describing anatomies. For biometry estimation, we annotated bounding boxes to identify regions of interest and used MedSAM \cite{ma2024segment} for mask label inference. Due to the poor quality of generated mask labels for the cerebellum and femur classes, manual reannotation was performed, resulting in 574, 43, 75 mask labels. Ground-truth biometric parameters were manually annotated by sonographers during the scanning process.

\textbf{Evaluation metrics.} In keyframe detection (stage 1), we used four metrics to evaluate the performance of predicted keyframes $p$ compared with the ground truth $y$, as detailed in Fig. \ref{fig:sup} of \textit{supplementary}. 1) \textit{`Cosine simi.'} and 2) \textit{`absolute time err.'} measure keyframe feature similarity and absolute timestamp error in a set-to-set manner. 
As some predicted keyframes exhibiting large time errors share the same appearance as the ground truth, 3) \textit{`correct time err.'}  measures the absolute keyframe timestamp error only when the maximum feature similarity is below a certain threshold (e.g., 0.96). 4) \textit{`Keyframe num. err.'} indicates the absolute error in the number of predicted keyframes. 
In the keyframe captioning (stage 2), two conventional metrics \textit{`BLEU'} and \textit{`ROUGE-L'} were utilized, following \cite{alsharid2022weakly}. For the final stage 3, absolute error and relative error of AC, HC, BPD, Cereb, FL biometry w.r.t. the clinical measurement were computed. 

\textbf{Implementation.} MMSummary is implemented with PyTorch on RTX 8000. In the pipeline, $f_{I}(\cdot)$, $f_{G}(\cdot)$, $f_{T}(\cdot)$ are frozen, while other parameters are updated with AdamW optimizer and a learning rate of 2e-5. $\tau$ and $\tau'$ are set as 0.96, 0.8.

\begin{table}[t!]
\centering
\caption{Quantitative experimental results for each of the three stages.}
\label{table:results}
\scalebox{0.65}{\begin{tabular}{p{1.2cm} | p{3.3cm} | c | p{0.9cm}  | p{0.9cm}  | p{0.9cm}  |  p{0.9cm} |  p{0.9cm} |  p{0.9cm} |  p{0.9cm} |  p{0.9cm} |  p{0.9cm} |  p{0.9cm} |  p{0.9cm} |  p{0.9cm} }
\toprule[1.2pt]

&\textbf{Scan time saved (\%)}&\textbf{Cosine simi.(\%) $\uparrow$}&\multicolumn{4}{c|}{\textbf{Absolute time err.(s) $\downarrow$}}&\multicolumn{4}{c|}{\textbf{Correct time err.(s) $\downarrow$}}&\multicolumn{4}{c}{\textbf{Keyframe num. err.$\downarrow$}} \\
\cline{2-15}

\multirow{7}*{\textbf{Stage 1}}&33.25 (35\%$\times$95\%)&94.12$~_{\pm 11.29}$&\multicolumn{4}{c|}{11.96$~_{\pm 5.18}$}&\multicolumn{4}{c|}{9.46 $~_{\pm 4.58}$}&\multicolumn{4}{c}{9$~_{\pm 5}$}   \\

&31.50 (35\%$\times$90\%)&94.19$~_{\pm 11.50}$&\multicolumn{4}{c|}{10.57$~_{\pm  4.81}$}&\multicolumn{4}{c|}{8.50$~_{\pm 4.53}$}&\multicolumn{4}{c}{6$~_{\pm 4}$}   \\

&26.25 (35\%$\times$75\%)&96.89$~_{\pm 3.75}$&\multicolumn{4}{c|}{\textbf{10.37$~_{\pm 6.29}$}}&\multicolumn{4}{c|}{7.98$~_{\pm 5.90}$}&\multicolumn{4}{c}{\textbf{4$~_{\pm 3}$}} \\

&17.50 (35\%$\times$50\%)&96.12$~_{\pm 2.76}$&\multicolumn{4}{c|}{11.36$~_{\pm 6.55}$}&\multicolumn{4}{c|}{8.05$~_{\pm 5.53}$}&\multicolumn{4}{c}{\textbf{4$~_{\pm 3}$}}  \\

&0 (35\%$\times$0\%)&\textbf{97.02$~_{\pm 2.74}$}&\multicolumn{4}{c|}{13.32$~_{\pm 6.15}$}&\multicolumn{4}{c|}{\textbf{7.66$~_{\pm 5.31}$}}&\multicolumn{4}{c}{5$~_{\pm 3}$}  \\

\cline{2-15}

&\cellcolor[gray]{.93} 0 (NQ/expert)
&\cellcolor[gray]{.93} {97.19/96.47}
&\multicolumn{4}{c|}{\cellcolor[gray]{.93} 13.61/12.00}
&\multicolumn{4}{c|}{\cellcolor[gray]{.93} {8.14/6.17}}
&\multicolumn{4}{c}{\cellcolor[gray]{.93} 5/4}  \\

&\cellcolor[gray]{.93} 0 (\textit{p}-value)
&\cellcolor[gray]{.93} 0.627 
&\multicolumn{4}{c|}{\cellcolor[gray]{.93} 0.629 }
&\multicolumn{4}{c|}{\cellcolor[gray]{.93} 0.505 }
&\multicolumn{4}{c}{\cellcolor[gray]{.93} 0.747 }   \\

\bottomrule[1.2pt]
\toprule[1.2pt]

&\textbf{Method}&\textbf{Bleu-1 (\%) $\uparrow$}&\multicolumn{3}{c|}{\textbf{Bleu-2 (\%) $\uparrow$}}&\multicolumn{3}{c|}{\textbf{Bleu-3 (\%) $\uparrow$}}&\multicolumn{3}{c|}{\textbf{Bleu-4 (\%) $\uparrow$}}&\multicolumn{3}{c}{\textbf{ROUGE-L (\%) $\uparrow$}}  \\
\cline{2-15}

\multirow{6}*{\textbf{Stage 2}}&Alsharid et. al \cite{alsharid2022weakly}&39.00$~_{\pm 38.18}$&\multicolumn{3}{c|}{19.22$~_{\pm 34.47}$}&\multicolumn{3}{c|}{8.94$~_{\pm 20.44}$}&\multicolumn{3}{c|}{5.02$~_{\pm 17.70}$}&\multicolumn{3}{c}{44.70$~_{\pm 39.15}$} \\

&ClipCap \cite{mokady2021clipcap}&75.80$~_{\pm 36.26}$&\multicolumn{3}{c|}{68.31$~_{\pm 40.01}$}&\multicolumn{3}{c|}{46.33$~_{\pm 43.87}$}&\multicolumn{3}{c|}{41.53$~_{\pm 42.75}$}&\multicolumn{3}{c}{80.17$~_{\pm 34.19}$} \\

&Ours w/ audio data&\textbf{80.77$~_{\pm  32.96}$}&\multicolumn{3}{c|}{\textbf{74.40$~_{\pm  37.19}$}}&\multicolumn{3}{c|}{\textbf{57.42$~_{\pm 42.42}$}}&\multicolumn{3}{c|}{\textbf{50.75$~_{\pm 42.99}$}}&\multicolumn{3}{c}{\textbf{85.31$~_{\pm 30.49}$}} \\

&Ours&79.04$~_{\pm 33.69}$&\multicolumn{3}{c|}{71.65$~_{\pm 37.66}$}&\multicolumn{3}{c|}{48.68  $~_{\pm 42.85}$}&\multicolumn{3}{c|}{45.40$~_{\pm 41.67}$}&\multicolumn{3}{c}{83.05$~_{\pm 31.04}$} \\

\cline{2-15}

&\cellcolor[gray]{.93} Ours (NQ/expert)
&\cellcolor[gray]{.93} {78.57/80.31}
&\multicolumn{3}{c|}{\cellcolor[gray]{.93} 71.08/73.20}
&\multicolumn{3}{c|}{\cellcolor[gray]{.93} 50.87/42.68}
&\multicolumn{3}{c|}{\cellcolor[gray]{.93} 45.23/45.86}
&\multicolumn{3}{c}{\cellcolor[gray]{.93} 82.70/83.99} \\

&\cellcolor[gray]{.93} Ours (\textit{p}-value)
&\cellcolor[gray]{.93} {0.626}
&\multicolumn{3}{c|}{\cellcolor[gray]{.93} 0.719}
&\multicolumn{3}{c|}{\cellcolor[gray]{.93} 0.255}
&\multicolumn{3}{c|}{\cellcolor[gray]{.93} 0.941}
&\multicolumn{3}{c}{\cellcolor[gray]{.93} 0.699} \\

\bottomrule[1.2pt]
\toprule[1.2pt]

&\textbf{Method}&\textbf{AC(mm) $\downarrow$}&\multicolumn{3}{c|}{\textbf{HC(mm) $\downarrow$}}&\multicolumn{3}{c|}{\textbf{BPD(mm) $\downarrow$}}&\multicolumn{3}{c|}{\textbf{Cereb(mm) $\downarrow$}}&\multicolumn{3}{c}{\textbf{FL(mm) $\downarrow$}} \\
\cline{2-15}
 
\multirow{7}*{\textbf{Stage 3}}&Reference (mask label)&4.23$~_{\pm  3.28 ~(2.82\%)}$&\multicolumn{3}{c|}{4.46$~_{\pm  2.72 ~(2.58\%)}$}&\multicolumn{3}{c|}{1.02$~_{\pm  0.67 ~(2.15\%)}$}&\multicolumn{3}{c|}{0.84$~_{\pm  0.87 ~(4.13\%)}$}&\multicolumn{3}{c}{0.57$~_{\pm  0.36 ~(1.83\%)}$}  \\

&Bano et. al \cite{bano2021autofb}&7.65$~_{\pm  5.84 ~(5.10\%)}$&\multicolumn{3}{c|}{6.77$~_{\pm  3.70 ~(3.93\%)}$}&\multicolumn{3}{c|}{2.74$~_{\pm  2.45 ~(5.81\%)}$}&\multicolumn{3}{c|}{3.22$~_{\pm  2.21 ~(15.89\%)}$}&\multicolumn{3}{c}{1.86$~_{\pm  1.89 ~(5.97\%)}$}  \\

&Nagabotu et. al \cite{nagabotu2023precise}&5.26$~_{\pm  3.96 ~(3.51\%)}$&\multicolumn{3}{c|}{7.38$~_{\pm  6.44 ~(4.28\%)}$}&\multicolumn{3}{c|}{1.97$~_{\pm  2.23 ~(4.18\%)}$}&\multicolumn{3}{c|}{3.40$~_{\pm  3.04 ~(16.77\%)}$}&\multicolumn{3}{c}{\textbf{0.95$~_{\pm  1.46 ~(3.06\%)}$}}  \\

&Ours w/ rand prompt&6.54$~_{\pm  4.84 ~(4.35\%)}$&\multicolumn{3}{c|}{8.46$~_{\pm  4.29 ~(4.90\%)}$}&\multicolumn{3}{c|}{2.42$~_{\pm  2.35 ~(5.13\%)}$}&\multicolumn{3}{c|}{1.82$~_{\pm  1.11 ~(8.98\%)}$}&\multicolumn{3}{c}{3.40$~_{\pm  2.96 ~(10.94\%)}$}  \\

&Ours
&\textbf{5.23$~_{\pm  3.81 ~(3.48\%)}$}
&\multicolumn{3}{c|}{\textbf{5.25$~_{\pm  3.44 ~(3.04\%)}$}}
&\multicolumn{3}{c|}{\textbf{1.25$~_{\pm  0.99 ~(2.65\%)}$}}
&\multicolumn{3}{c|}{\textbf{1.05$~_{\pm  1.10 ~(5.16\%)}$}}
&\multicolumn{3}{c}{2.78$~_{\pm  2.64 ~(8.99\%)}$}  \\

\cline{2-15}
 
&\cellcolor[gray]{.93} Ours (NQ/expert)
&\cellcolor[gray]{.93} 5.57/4.50
&\multicolumn{3}{c|}{\cellcolor[gray]{.93} 4.77/6.20}
&\multicolumn{3}{c|}{\cellcolor[gray]{.93} 1.27/1.22}
&\multicolumn{3}{c|}{\cellcolor[gray]{.93} 1.15/0.86}
&\multicolumn{3}{c}{\cellcolor[gray]{.93} 2.38/3.98}  \\

&\cellcolor[gray]{.93} Ours (\textit{p}-value)
&\cellcolor[gray]{.93} 0.972
&\multicolumn{3}{c|}{\cellcolor[gray]{.93} 0.214}
&\multicolumn{3}{c|}{\cellcolor[gray]{.93} 0.186}
&\multicolumn{3}{c|}{\cellcolor[gray]{.93} 0.227}
&\multicolumn{3}{c}{\cellcolor[gray]{.93} 0.740}  \\

\bottomrule[1.2pt]

\end{tabular}}
\end{table}

\textbf{Quantitative results.} According to the frame-frame similarity matrix $S$, frames exhibiting a high similarity score above 0.97 compared to any keyframes within the video account for approximately 35\% of the entire video. For these frames, we observed that the sonographer was typically performing fine-grained probe adjustments, interpretations, or biometry. By randomly dropping frames during these activities (i.e., from these 35\% frames) in various proportions (i.e., 0\%, 50\%, 75\%, 90\%, or 95\%), we obtained several shortened videos. We then fed these videos into the keyframe detection model learned in \textbf{stage 1} for evaluation. The shortened video, with less frame redundancy, was found to simplify the keyframe detection process, leading to the lowest \textit{keyframe num. err.} when dropping 50\%, 75\% frames, as in Table \ref{table:results}. Moreover, our results demonstrate that keyframe detection maintains a good performance with only a 0.84s increase in \textit{correct time err.}, even for the case of saving 31.5\% scanning time. This indicates the potential of MMSummary to significantly reduce the time required for fetal US examinations without compromising keyframe detection accuracy. 

In \textbf{stage 2}, our method outperforms the fetal US image captioning method \cite{alsharid2022weakly} and baseline \cite{mokady2021clipcap} with increments of 40.04\%, 3.24\% in Bleu-1, respectively. Owing to the flexibility and reasoning capability of the language model, the sonographer's speech transcribed from audio data using \cite{bain2023whisperx} can also be incorporated as an additional input prompt to GPT2 for caption reasoning. After fine-tuning GPT2 for 10 epochs, our model achieved 5.35\% improvement in Bleu-4. 

For \textbf{stage 3}, compared with fetal biometry estimation methods \cite{bano2021autofb,nagabotu2023precise}, our method obtained the best performance in AC, HC, BPD, Cereb biometry. Moreover, we also investigated replacing the textual prompt with a random prompt that lacked any prior information. This negatively impacted the biometry estimation performance, demonstrating the importance of semantic relations embedded in textual prompts to enhance biometry estimation performance.

Finally, we report quantitative results of our method on US videos acquired by sonographers with two skill levels, namely newly qualified (NQ) and expert, shown in gray cells of Table \ref{table:results}. An independent two-tailed t-test was performed to compute \textit{p}-values. The analysis indicates that there is no significant difference (\textit{p}-value$>$0.1) in the performance of our method on NQ and expert videos across all metrics, suggesting that our method is robust to different operator expertise.

\begin{figure}[t]
\centering
\includegraphics[width=1.0\textwidth]{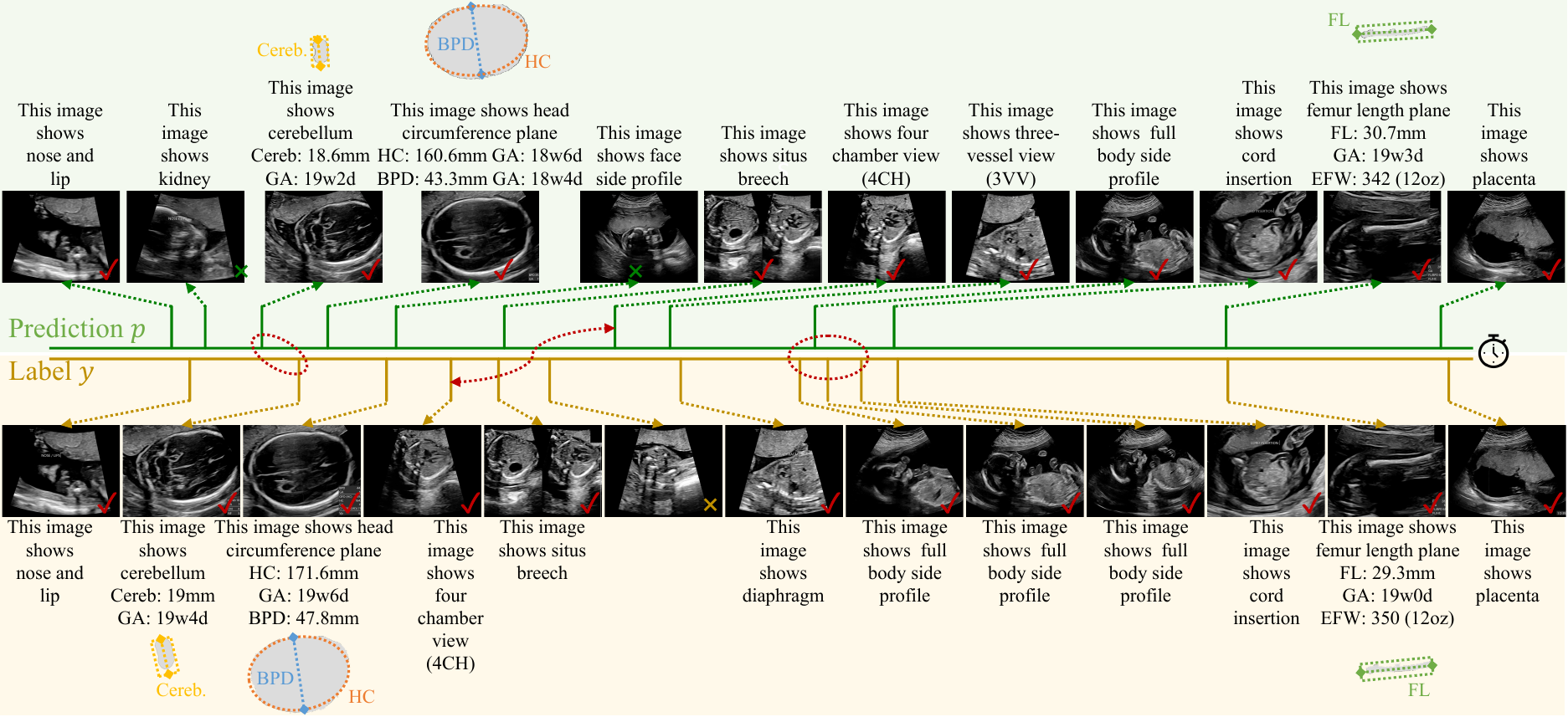}
\caption{MMSummary prediction (upper part) and ground truth (lower part).} \label{fig:result}
\end{figure}

\textbf{Qualitative results.} Fig. \ref{fig:result} (upper part) presents a multimodal summary of an example video generated by our pipeline. Compared with the ground truth (lower part), most keyframes are correctly detected. However, some predicted keyframes exhibit large time errors (left two red dotted indicators). Additionally, our method only detects one `full body side profile' keyframe (right red dotted circle), while three similar frames are shown in the ground truth. Despite these discrepancies, the accuracy of the multimodal summary is not affected, as the image content of the keyframe is still correct. The predicted keyframes marked with green crosses are redundant or non-informative. In practice, these would be easily removed by a sonographer. The keyframe with a yellow cross is present in the ground truth but not predicted. This may be due to a sonographer incorrectly recording the screenshot since it is of poor image quality. The predicted captions and biometry estimations are reasonably consistent with the ground truth. Captioning errors mainly arise for keyframes with low image quality and in distinguishing cardiac views, which are visually similar in appearance. 

\section{Conclusion}

This paper focuses on the automated generation of a multimodal summary for a given medical video, using a fetal US video as a potential clinical use case. Imitating the real examination process, we propose a three-stage MMSummary system that automatically generates a concise subset of keyframes summarizing the useful video content, accompanied by corresponding text descriptions and biometry estimations, for a fetal US video. This system has the potential to save fetal US scanning time and reduce the expertise needed to scan. 

\begin{credits}
\subsubsection{\ackname} The authors acknowledge UKRI grant reference (EP/X040186/1), EPSRC grant (EP/T028572/1), and ERC grant (ERC-ADG-2015 694581, project PULSE). We thank Prof. Andrew Zisserman, Dr. Tengda Han, and Jayne Lander for sharing valuable comments.

\subsubsection{\discintname}
The authors have no competing interests to declare that are
relevant to the content of this article. 

% Or: Author A has received research
% grants from Company W. Author B has received a speaker honorarium from
% Company X and owns stock in Company Y. Author C is a member of committee Z.
\end{credits}

\bibliographystyle{splncs04}
\bibliography{Paper-0399.bib}

\newpage

\section{Supplementary}

\begin{figure}[h!]
\centering
\includegraphics[width=\textwidth]{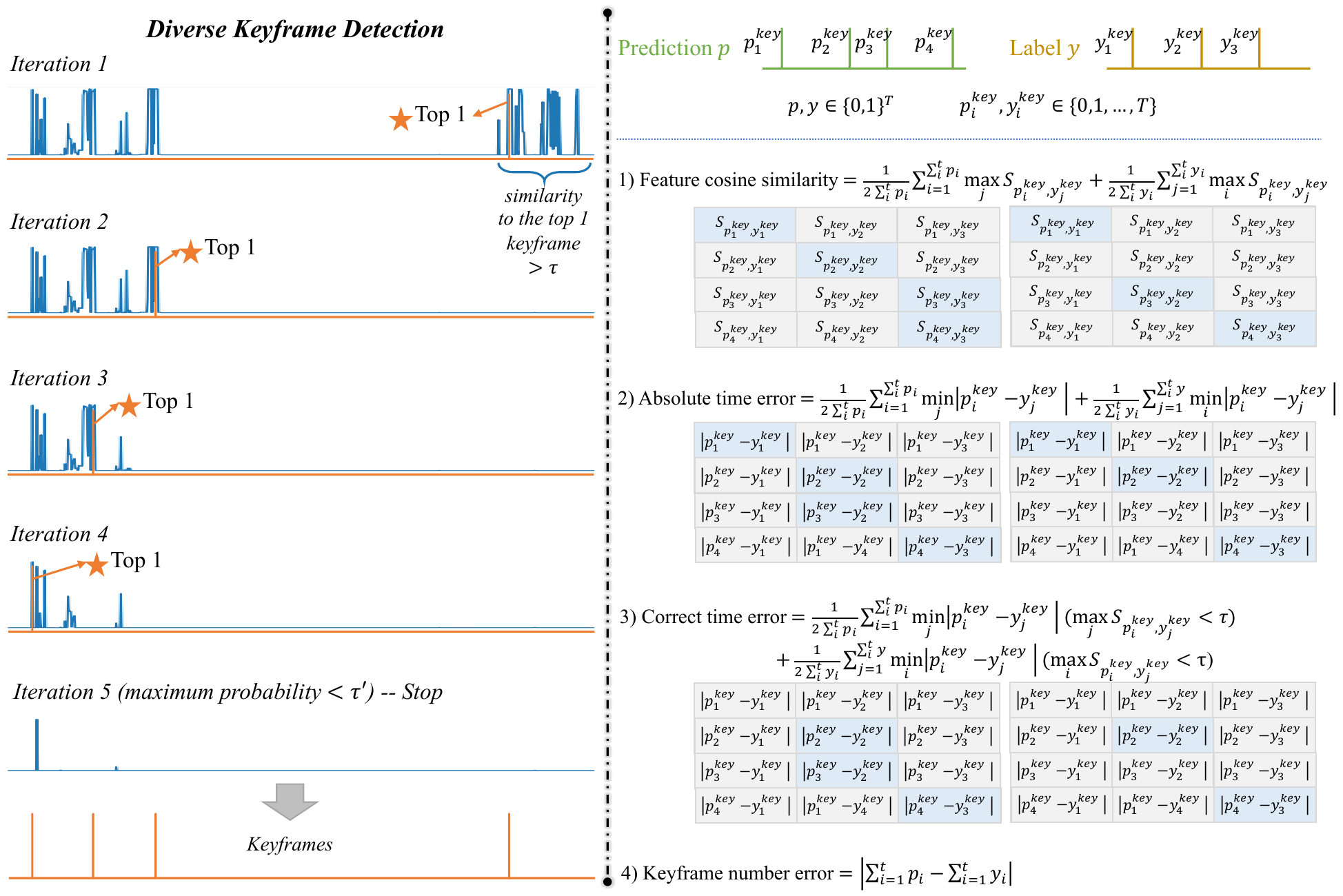}
\caption{Left: diverse keyframe detection algorithm. Right: evaluation metrics in the first keyframe detection stage.} 
\label{fig:sup}
\end{figure}

\begin{figure}[htp!]
\centering
\includegraphics[width=\textwidth]{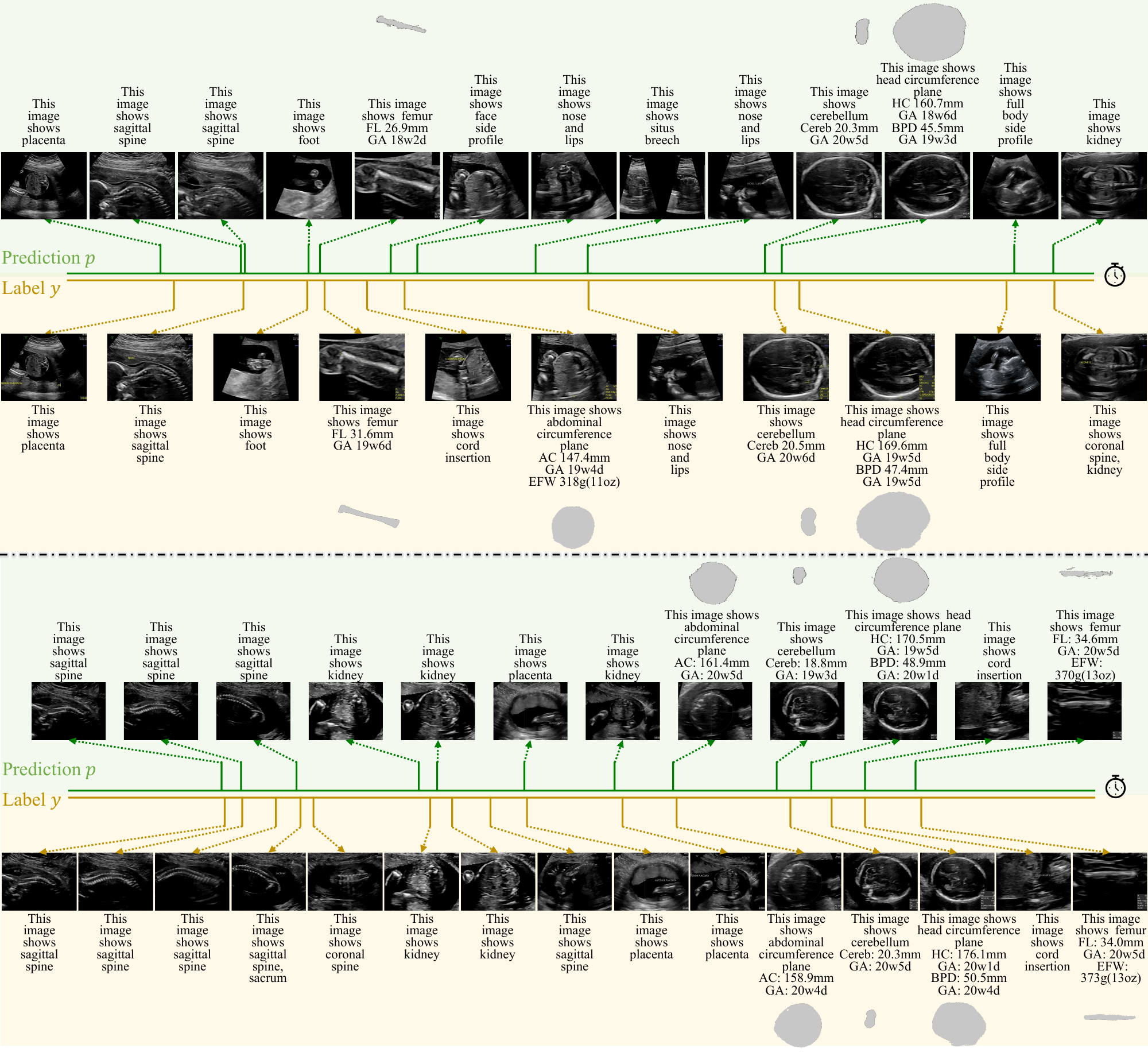}
\caption{More qualitative comparison of multimodal summarization between our prediction and ground truth.} 
\label{fig:result_sup}
\end{figure}

\end{document}